\pdfoutput=1
\documentclass[letterpaper, 10 pt, conference, hyphens]{ieeeconf}  

\IEEEoverridecommandlockouts                              
\usepackage{booktabs}
\usepackage[utf8]{inputenc}
\usepackage[T1]{fontenc}
\usepackage{amsmath}
\usepackage{amssymb}
\usepackage{bm}
\usepackage{epsfig}
\usepackage{graphics}
\usepackage{mathptmx}
\usepackage{hyperref}
\usepackage{mathtools}
\usepackage{xcolor}

\title{\LARGE \bf A Multi-Modal Wildfire Prediction and Personalized Early-Warning System Based on a Novel Machine Learning Framework
}

\author{Rohan Tan Bhowmik$^{1}$
\thanks{$^{1}$ Stanford University School of Medicine, Stanford, California and The Harker School, San Jose, California; rbhowmik@stanford.edu and 23rohanb@students.harker.org}
}

\begin{document}

\maketitle
\thispagestyle{empty}
\pagestyle{empty}

\begin{abstract}

Wildfires are increasingly impacting the environment, human health and safety at a global level. Among the top 20 California wildfires, those in 2020-2021 burned more acres than the last century combined. California’s 2018 wildfire season caused economic damages of \$148.5 billion. Among millions of people impacted each year, those living with disabilities ($\sim$15\% of the world population) are disproportionately impacted due to inadequate means of early alert. In this project, a multi-modal wildfire prediction and personalized early warning system has been developed based on an advanced machine learning architecture. Environmental and meteorological sensor data from the Environmental Protection Agency and historical wildfire data in 1.4 miles x 1.4 miles square grids from 2012 to 2018 have been compiled to establish a comprehensive wildfire database, the largest of its kind. Next, a novel U-Convolutional-LSTM (Long Short-Term Memory) neural network was designed and implemented with a special architecture  designed to extract key spatial and temporal features from environmental parameters inherent within contiguous weather data that are indicative of impending wildfires. Environmental and meteorological factors were incorporated into the database and classified as leading indicators, correlated to higher risks of wildfire conception, and trailing indicators, correlated to smaller wildfires. Additionally, geological data was used to provide better wildfire risk assessment. This novel spatio-temporal neural network achieved >97\% accuracy, as opposed to $\sim$76\% using traditional convolutional neural networks, successfully predicting 2018’s five most devastating wildfires 5-14 days in advance. Finally, a personalized early warning system, tailored to individuals with sensory disabilities or respiratory exacerbation conditions, was proposed. This technique would enable fire departments to anticipate and prevent wildfires before they strike and provide early warnings for at-risk individuals for better preparation, thereby saving lives and reducing economic damages.
\end{abstract}

\section{\textbf{INTRODUCTION}}

In recent years, wildfires are progressively posing a global threat to the infrastructure, economy, and quality of life around the world. Within the United States alone, the annual average of acreage burned by wildfires since 2020 is double the acreage burned during the 1990s [1].  Specifically in California, over 4.2\% of the land area was burned by the 2020 wildfire season alone [2].

Severe environmental circumstances are increasingly being induced by rapid climate changes, such as intense drought conditions and extreme temperatures [3, 4]. These weather conditions are being caused by abnormal precipitation cycles, with acute lack of rain and snow in California resulting in abnormally dry environment. Resulting shortages of moisture within the soil, air, and vegetation remove any water content that would otherwise act as a buffer against fire spread, turning land impacted by droughts into a tinderbox, creating ideal conditions for wildfires sparks to ablaze and exacerbate into unprecedented scales of destruction. These conditions are increasingly enabling wildfires to burn out of control, turning into major conflagrations before the wildfire management teams and firefighters can intervene and properly manage, mitigate, and control the burns.

Additionally, forest overgrowth in recent years and lack of controlled burns led to extra foliage to burn through, which has exacerbated the extraordinarily devastating recent wildfire seasons [5]. The common fire suppression strategy from wildfire management services, such as the U. S. Forest Service, is to immediately control and put out wildfire upon discovery. This, however, leads to a gradual accumulation of unburnt foliage that should have been eliminated naturally by wildfires, building up a large amount of combustible fuel for future fires. Burning through decades worth of additional forestry has allowed recent wildfires to spread uncontrollably and shatter records year after year.

Instances of particularly large wildfires, such as those burning areas over 10,000 hectares (around 25,000 acres) have been documented throughout California's history. For example, a large influx of people settling into California led to a large frequency of wildfires with anthropogenic origins. This wave of conflagrations peaked in the 1920s decade, leading to over 4.2 million acres of land to be burned through wildfires [6]. However, its severity dwindles in comparison to the 2020 wildfire season, which alone burned through 4.3 million acres of land [2].

This has massive implications on the economy, with the 2020 wildfire season costing California almost \$4.3 billion in property damage and fire suppression resources [2]. This owes to the rapid shift of demographics in California; while the 1920s wildfires did burn through significant portions of land, a smaller population and number of structures means little to no fatalities and minimal property damage. With a massive proliferation of industrialization allowing civilization to settle throughout California, any wildfire of large magnitude would increasingly cause serious issues towards infrastructure, human health, and quality of life [6]. On the other hand, increased resources and new technologies for preemptive wildfire suppression would allow for faster wildfire management and damage mitigation. As such, the ability to predict wildfire location and severity would be a significant breakthrough for preventative measures in stopping small flames from transgressing into larger conflagrations and control wildfire spread smartly.

This project explores a multi-modal wildfire prediction and personalized early warning system based on a novel machine learning framework towards aiding the suppression of major fires before they burn out of control. Historical environmental, meteorological, and geological data along with past wildfire data were gathered to create a database, the largest of its kind, which was used to train the machine learning network with a unique architecture specifically suited for wildfire prediction based on learning the spatio-temporal patterns inherent in the leading and trailing indicators. The model outputs risks of how likely a major wildfire would ignite in a certain area within the near future, an assessment that would be useful for both diverting wildfire management resources early and for civilians to begin preparing for evacuations. Instead of attempting to mitigate smaller wildfires necessary for keeping forest growth in check and preventing wildfire crises in the future, the objective is to prevent these smaller wildfires from exploding into devastating conflagrations by predicting such likelihoods and taking preemptive measures, which would save money, infrastructure, and lives.

\section{\textbf{PRIOR RESEARCH}}

While there has been extensive ongoing research in this domain, previous work primarily focused on detection, rather than prediction of wildfires [7-12]. Many of these solutions require a large number of resources to the point where the cost of deployment may outweigh perceived benefits. Existing prediction techniques are also limited in prediction scope (area analyzed) and accuracy.

The topic of early warning systems for natural disasters in general, including wildfire, is widely examined in both scientific research and governmental policies. Current early warning systems offer two methods of mitigating losses from natural disasters [8, 9]. The first method involves predicting the likelihood of a natural disaster occurring in the near future within a region by comparing its meteorological and geographic conditions to those recorded preceding historical disasters. The second method is based on detecting the onset of an imminent disaster by observing definite precursors and signs for these geological events. Different types of natural disasters benefit differently from each early warning method. Disasters whose severity progresses over time, such as wildfires, tend to show noticeable indicators and/or follow predictable trends, a task suited for a predictive system that could alert, warn, and advise evacuation of the potentially impacted populations.

Researchers have also created methods of evaluating the cost-benefits of implementing early warning systems. As the maintenance costs from deployment, regular quality inspections, and sunk costs from false alarms and inaccurate reporting go down, overall benefit from the system goes up [9]. Thus, recently proposed early warning systems utilize modern technology to create more efficient and accurate methods of alerting to natural disasters [10]. Notably, the sharp spike in the number of active satellites due to their increase in widespread viability has enabled continuous monitoring of the entire globe. Different types of sensors placed on satellite networks create a broad, aerial analysis of surface temperature, water elevation, and vegetation as well as means for detecting potential natural disasters.

Another form of early warning system is the emerging remote sensor network. Similar to satellites, wireless, land-situated sensors spread throughout regions of interest relay readings to central stations, which then process recorded data to predict or detect natural hazards. Comparatively, remote sensor networks are cheaper to deploy, higher in precision, but narrower in scope than satellite networks [11, 12].

\begin{figure*}[thpb]
      \centering
      \label{figure-previous}
      \includegraphics[width = 6in]{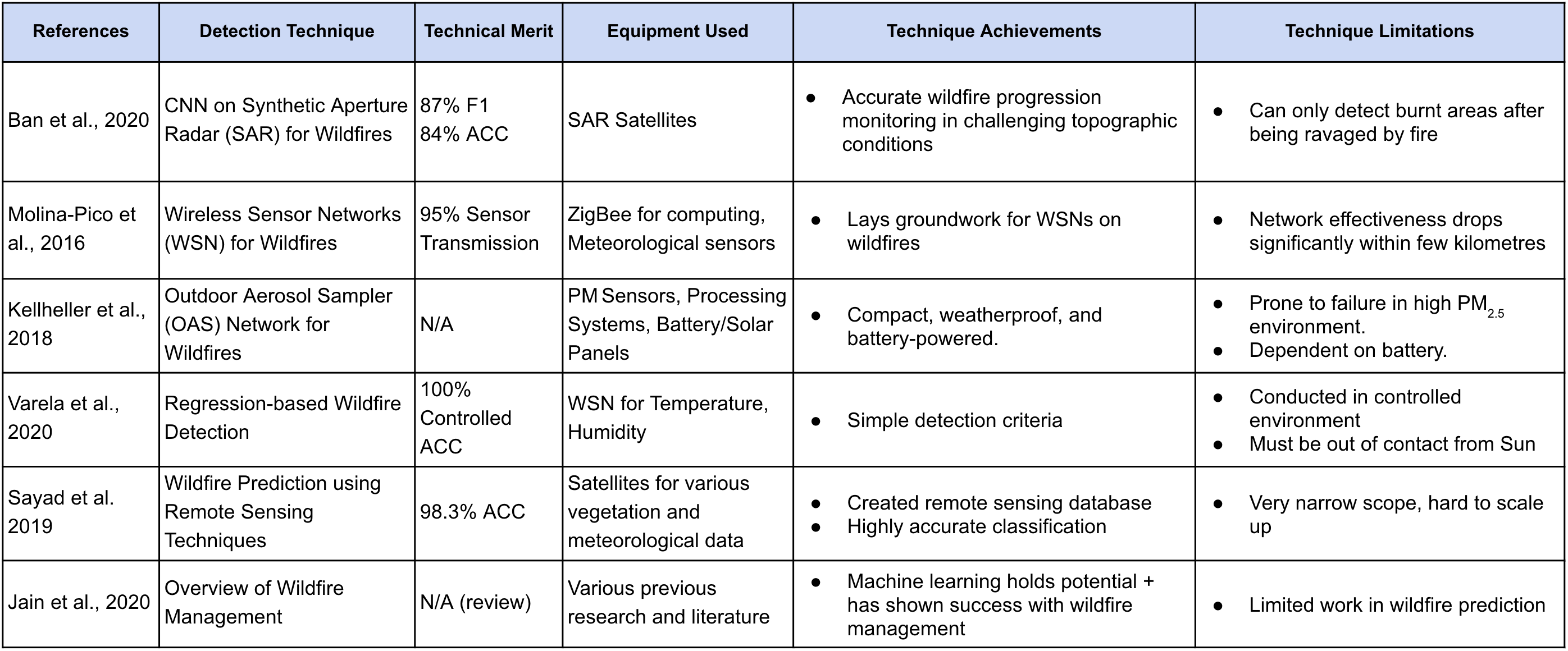}
      \caption{A summary of previous research conducted on wildfire analysis, detection, and prediction, detailing each technique's achievements and limitations (F1 = Harmonic Mean of Specificity and Sensitivity; ACC = Accuracy). Specifically, these techniques reveal a lack of a solution that i) accurately predicts large wildfires, ii) utilizes key, consistent wildfire indicators, and iii) allow for real-time applications with readily available data.
}
\end{figure*}

\hyperref[figure-previous]{Figure 1} provides a summary of previous research conducted on wildfire analysis, detection, and prediction, detailing each technique's major achievements as well as key limitations.

Jain et al. provides an overview of wildfire management techniques, especially for Artificial Intelligence (AI) techniques tackling wildfire issues [7]. Overall, machine learning techniques hold potential for wildfire applications, yet this is an emerging field with recent studies exploring new directions. Ban et al. created a wildfire progression prediction tool using a Convolutional Neural Network (CNN) trained on Synthetic Apeture Radar (SAR) data [8]. Although the technique showed high accuracy in tracking wildfire progression regardless of geological conditions that would make the prediction complicated, the technique can only predict wildfires after they have already burnt large portions of land. Molina-Pico et al. use wireless sensor networks for monitoring basic meteorological factors for wildfires, setting the foundation for sensor networks in this field of study [9]. However, this technique requires a dense network of sensors, as its accuracy decrases after traveling a few kilometers. Kellheller et al. used an Outdoor Aerosol Sampler (OAS) network, which measures environmental levels in the air, for monitoring wildfire spread remotely [10]. The OAS devices are durable and the spatially efficient nature make it effective for wildfire monitoring, but its battery dependence and inaccurate extreme particulate matter reading holds the technique back. Varela et al. created a regression model for detecting wildfires, which was able to achieve impressive accuracy in a controlled environment using only a simple detection criteria [11]. Since it must be in a controlled environment, however, it doesn't perform as well in the real world such as under sun exposure. Sayad et al. created a wildfire prediction technique using satellite data for geological and meteorological factors [12]. While their technique is very accurate, it focuses on a localized area for analysis which is difficult to scale up for larger impact.

Overall, these techniques reveal a lack of a solution that i) predicts wildfires with high accuracy, ii) utilizes key, consistent leading and trailing indicators of wildfires, and iii) allows for real-time applications with readily available data.

\section{\textbf{METHODS}}

This project has three main components: i) a large dataset consisting of historical wildfires as well as key leading and trailing indicators, ii) a novel spatio-temporal machine learning architecture for wildfire prediction, and iii) a personalized early warning system. The wildfire database, the largest of its kind, contains the location and severity of historical wildfires as well as environmental, meteorological, and geological data that encode indicative probabilities of future wildfires. The AI model is trained to take in the aforementioned data to provide wildfire risk assessments for locations across California, with its novel architecture created to process and leverage the unique data characteristics. The early warning system utilizes this wildfire risk assessment to provide warnings and advise evacuation preparation measures, with different specifications in alert criteria and content for different populations.

\subsection{A Comprehensive Wildfire Dataset}

The wildfire prediction system consists of a unique spatio-temporal neural network architecture that was trained on historical environmental and meteorological data across California. Specifically, the objective of the training process is to teach the neural network which patterns encoded within the historical data correspond to a certain wildfire's location and severity, which would allow it to predict future wildfires based on the current data from a network of sensors.

\subsubsection{Leading and Trailing Indicators}

Two categories of environmental and meteorological factors can be derived from identifying the key causes and effects of wildfires: leading indicators and trailing indicators. 

Leading indicators demonstrate particular trends in areas where sparks or lightning could produce flames. Certain patterns in environmental factors in an area produce conditions where a spark would be prone to enrage into a wildfire rather than be smothered and extinguished. Additionally, optimal weather conditions would help billow and fuel a small fire into a raging conflagration that would yield a much larger acreage and potential damage. Hence, leading indicators lead to the conceptions of wildfires. The leading indicators collected in this database included:

\begin{enumerate}
  \item Temperature (measured in $^{\circ}$C), which indicates conditions such as precipitation, water evaporation, etc., that would create an ideal environment for wildfires.
  \item Dew point (measured in $^{\circ}$C), the temperature to which air must be cooled to be fully saturated with water vapor. Dew point can be used to find relative humidity which indicates the dryness and thus inflammability of the area.
  \item Wind speed (measured in knots$/$hour), which would provide supply of air to fuel the fire and a means of carrying the burning soot to spread fire to its surroundings.
\end{enumerate}

On the other hand, trailing indicators show abnormal trends before a small flame transgresses into a large wildfire. After a fire begins to burn, warm, smoky air from the fire rises into the atmosphere and is carried into its surrounding. As compared to fires which are dependent on ideal landscape and environmental conditions to propagate, smoke is able to diffuse quickly through the air with or without the aid of a blowing wind. Nearby sensors which actively measure trailing indicators carried through the smoke can determine the existence and severity of an otherwise undetected wildfire. Hence, trailing indicators trail after the conception of wildfires. The trailing indicators, which often correlate with smoke, collected in this database included:

\begin{enumerate}
  \item PM$_{2.5}$ (measured in $\mu g/m^3$), the concentration of particulate matter in the air with diameter of $2.5 \mu m$ or less [13].
  \item PM$_{10}$ (measured in $\mu g/m^3$), the concentration of particulate matter in the air with diameter of $10 \mu m$ or less [13].
  \item CO or Carbon Monoxide levels (measured in parts per million, or ppm) [14].
  \item NO$_2$ or Nitrogen Dioxide levels (measured in parts per billion, or ppb) [15].
\end{enumerate}

\begin{figure*}[thpb]
      \centering
      \label{figure-camp}
      \includegraphics[width = 6in]{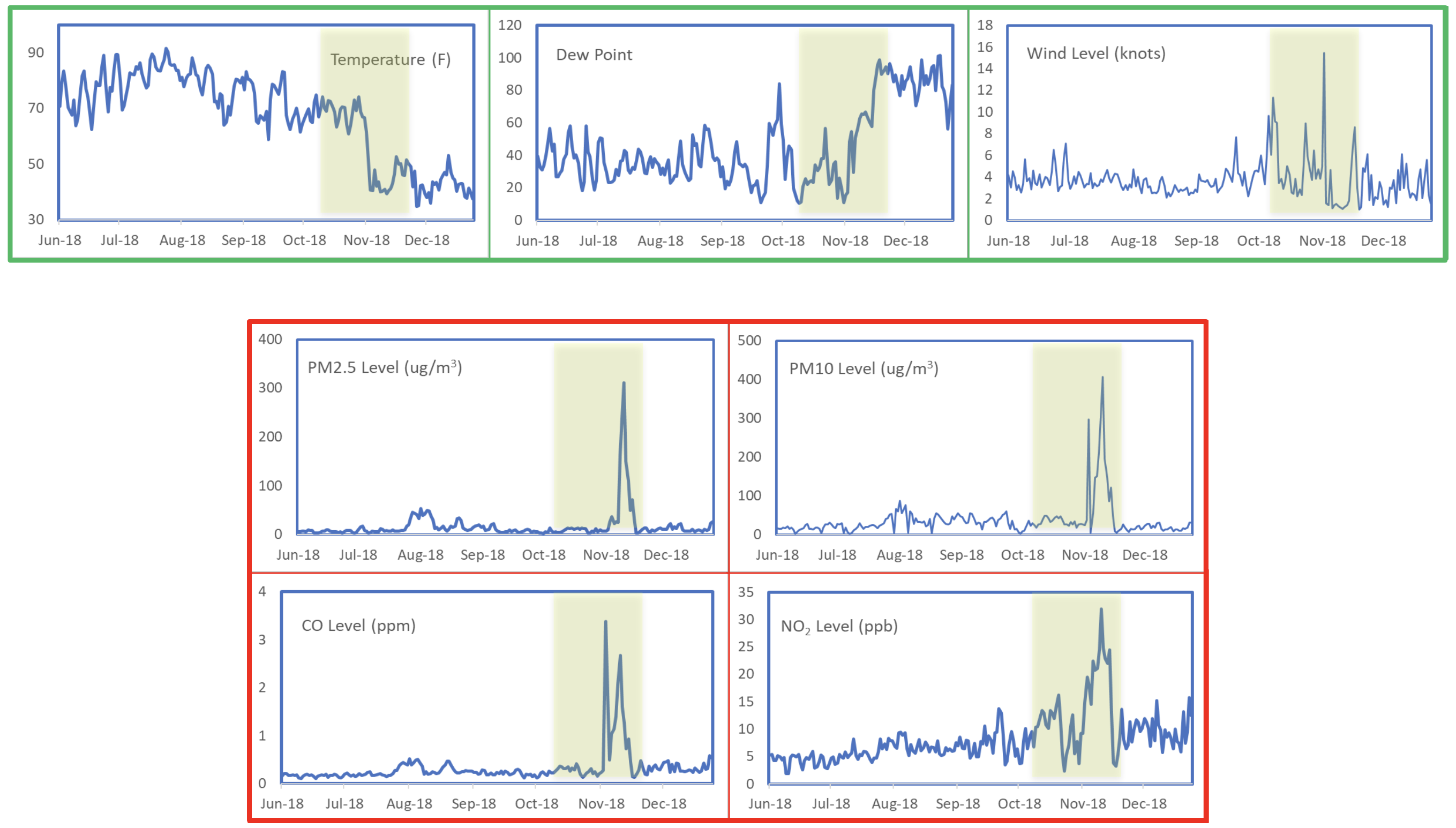}
      \caption{Graphs of leading indicators including temperature, dew point, wind speed (outlined in green) and trailing indicators including PM$_{2.5}$, PM$_{10}$, CO, and NO$_{2}$ concentrations (outlined in red) from June to December of 2018 over Paradise, California. The highlighted regions indicate the devastating Camp Fire's period, within which the graphs reflect abnormal trends right before and during the fire's conception. These trends provide the basis of predicting the risk of a wildfire igniting using leading and trailing indicator patterns.}
\end{figure*}

On November 8, 2018, a faulty powerline sparked an ignition which would eventually become the deadly Camp Fire in California, the single most expensive natural disaster globally in 2018 [16]. \hyperref[figure-camp]{Figure 2} shows seven graphs depicting each of the leading and trailing indicator readings over time for the city of Paradise, California, which was completely decimated by the notorious wildfire and led to the Pacific Gas and Electric Company (PG\&E) filing for bankruptcy. The graphs show how each measurement demonstrated abnormal readings and trends before and/or during the Camp Fire as highlighted in yellow-green (which covers late October to early December with the fire lasting from November 8 through 25). These trends provide the basis of predicting the risk of potential wildfires via automatic analysis of the leading and trailing indicator patterns using machine learning techniques.

\subsubsection{Environmental and Meteorological Data Processing}

Each of these seven leading and trailing indicators are available on the historical database of the Environmental Protection Agency (EPA) web page [17]. The data for each environmental and meteorological factor are partitioned into one-year files, each with daily readings from weather sensors across the United States.

After downloading data for the leading and trailing indicators, the daily readings were plotted onto a 2-dimensional plot of California as points, with the location and value of each point representing the location and reading of the corresponding weather sensor. Then, the data between the points were interpolated to create a predicted environmental/meteorological factor map, thereby turning a series of readings into an image. Additionally, all the data was transformed through a logarithmic function defined by $y = \log_2(x + 1)$. This was done to normalize the extreme readings during wildfires, where station's weather measurements could increase by more than 50 times from its normal levels.

\subsubsection{Geological Data Processing}

Additionally, land characteristics, including terrain elevation and local vegetation productivity, are also important in dictating the likelihood of a wildfire catching ablaze as well as the potential of a small fire to spread into a much larger one. As such, the location where the model makes a prediction depends on the geological specifications of the area under consideration.

Terrain elevation dictates much of the environmental and atmospheric processes occurring within the area [18]. For example, at rather high elevations, a lower level of oxygen means that fires will have lower likelihood of seriously exacerbating. Additionally, higher altitudes usually signify a sloped or mountainous area, which could mean a lusher and moister region of land on the windward side that is less likely to catch serious fires, or a drier and more combustible landscape on the leeward side's rain shadow that is more likely to catch serious fires. With all this considered, there is generally a smaller population density towards regions of higher elevation, meaning that while a fire would cause less damage to infrastructure and disruption to or loss of human life, it could also mean a longer response time to control the fire leading to more severe air pollution that would affect respiratory health of more sensitive groups. Similar considerations come into play with other elevation patterns and the terrain formations they represent, which are all key information that the machine learning model can use to train.

Vegetation productivity, a measurement reflected by the Normalized Difference Vegetation Index (NDVI), is an indicator of, given the absorbed influx of solar energy into a certain area, how much infrared energy is radiated [19]. Specifically, the higher the NDVI index, the more productive the vegetation, and the lower the NDVI index, the less productive the vegetation. For more healthy and productive plants, the more infrared light is emitted by working chlorophyll cells, and vice versa. Plant health and productivity can indicate the conditions of the terrain surrounding the plant: unhealthy plants could indicate an unusually dry or hot environment for the plant, which could indicate a greater volatility for large fires. Additionally, a lower NDVI index could indicate dying or dead foliage, the perfect fuel source for a conflagration to burn through.

\subsubsection{Wildfire Data Processing}

Historical wildfire data in California was collected from WildfireDB, an open-source dataset maintained by Vanderbilt University, UC Riverside, and Stanford University [20]. The dataset contains over 17 million data points with daily wildfire details from 2012 to 2018. Specifically, the wildfire activity is quantified through Fire Radiative Power (FRP), which is a measure of radiated infrared energy per unit of time that reflects the rate of burning biomass. This data was normalized through a fourth-root scale defined by $y=\sqrt[\leftroot{-2}\uproot{2}4]{x}$ to express fire activity as a function of temperature according to the Stefan-Boltzmann law (which states that $P \propto T^4$). This scaling ensures that smaller fires with near-negligible FRP measurements are not ignored.

In order to be utilized for wildfire prediction using the neural network framework described in the next section, the dataset needs to be transformed to form two-dimensional images with the x and y position of a pixel corresponding to a location over California and the pixel value representing the measured level of a certain indicator at that specific location. As these indicators largely follow cyclic trends that revolve around the human circadian rhythm, each pixel's value would then be the average measured indicator level at that location across the entire day. As this would be repeated for each day, the total data for a certain indicator would be a sequence of images in chronological order with respect to the date of the collected data. Since there are multiple indicators, the database consists of multiple such sequences of images.

Although time does add a third dimension to the sequence of two-dimensional images, the database cannot be handled as a collection of three-dimensional images. This is because while images contain patterns within the spatial domain, time-related data contains patterns within the temporal domain, meaning that both aspects of the database must be analyzed together during wildfire prediction. This insight led to the development of a unique machine learning architecture as narrated below.

\subsection{The Spatio-Temporal U-Convolutional-LSTM Network}

\begin{figure*}[thpb]
      \centering
      \label{figure-network}
      \includegraphics[width = 6in]{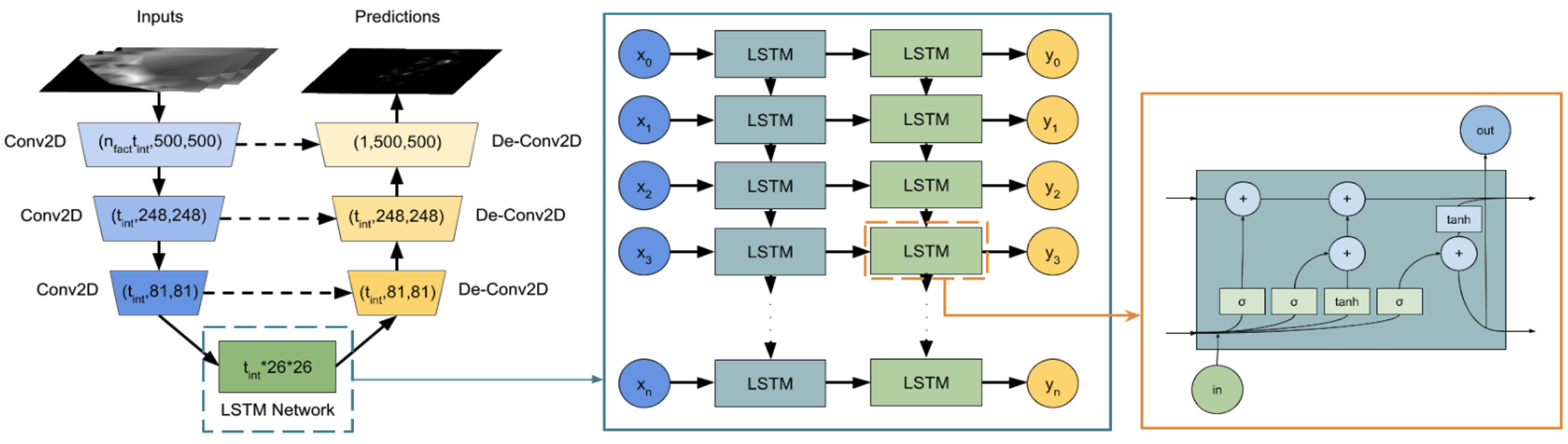}
      \caption{Architecture of the U-Convolutional-LSTM neural network. The left diagram reflects the U shaped network framework analogous to that of the U-Net, where the dashed arrows only exist to indicate symmetry of the U-Net by comparing analagous parts. The middle diagram details the 2-layer LSTM network found at the center of the ULSTM network. The right diagram details the LSTM unit that comprises the LSTM network.}
\end{figure*}

To effectively analyze the spatial and time-series data of the database, a novel spatio-temporal U-Convolutional-LSTM (Long Short-Term Memory) neural network, whose name references its architecture and purpose, was created. The structure of this unique neural network framework is depicted in \hyperref[figure-network]{Figure 3}. Spatio-temporal refers to the model's ability to effectively process and analyze data using both spatial and temporal feature extraction capabilities. Information in the spatial domain includes data represented as arrays and matrices. As such, spatial analysis is important in image-related tasks such as picture classification, object recognition, and image generation. On the other hand, information within the temporal domain includes data represented in sequences and series. As such, temporal analysis is important in chronological time-series data such as text analysis, music generation, and video manipulation. Hybrid and deeply meshed spatio-temporal neural network architectures have been shown to be effective in analyzing complex datasets where key information is encoded in both spatial and temporal dimensions concurrently [21 - 25].

The first fundamental aspect of the proposed U-Convolutional-LSTM architecture, which will be shortened to ULSTM for the sake of convenience, is the U-Convolutional network, or U-Net, architecture. The U-Net framework was proposed by Ronneberger, et al., in 2015, for biomedical imaging applications [26], which built on the fully-convolutional network specifically for semantic segmentation reported by Shelhamer, et al., in 2014 [27]. The first half of a U-Net consists of a series of convolutional layers, each of which transforms the input image into a smaller, compressed image containing the extracted features. Specifically, the convolutional layers pass a 2-dimensional filter over the input image, returning an output matrix with every cell equal to the filter and the region of the input image it passes over. At the end of the series of convolutional layers, a compact matrix of key spatial features is obtained. The second half of the network is structurally symmetric to the first half of the network in a mirror-image fashion, consisting of a series of deconvolutional layers. These deconvolutional layers perform the same operations that a convolutional layer does but with uniform padding that increases the dimensions of the input image. As such, while convolutional layers compress an input image, deconvolutional layers expand an input image. Overall, the U-Net transforms an input image to a condensed feature matrix, which contains key spatial features which the U-Net uses to form another image. Essentially, the U-Net is a generative network that specifically uses an input image to generate an output image of the same dimensions, but consisting of the extracted features. 

The second fundamental aspect of the proposed ULSTM architecture is the Long Short-Term Memory network, shortened to LSTM network. Originally proposed by Hochreiter, et al., and subsequently refined by many researchers, the LSTM network is a kind of recurrent neural network that specializes in temporal feature analysis [28]. By selectively learning long-term dependencies from previous inputs, LSTM networks recall stored input sequence information to synthesize outputs. As such, the LSTM network extracts key temporal features from a chronological list of inputs and processes them to form outputs. This is especially helpful for data that are time-sequential, such as consecutive frames of a video or the words of a sentence.

As illustrated in \hyperref[figure-network]{Figure 3}, the ULSTM network developed in this project is a synthesis of the U-Net and LSTM network's structures. Unlike either of those two networks, however, the ULSTM takes in a series of images as its input. Specifically, the ULSTM architecture processes and analyzes the previous two week's environmental and meteorological factor maps to output a risk map that details the probability of a fire starting within a day in each location over California. The overall architecture of the ULSTM network is analagous to that of the U-Net, with its first half being a series of convolutional layers that compresses the input images into a series of feature matrices. These feature matrices are then propogated through a 2-layer LSTM network which analyzes them while retaining short-term memory of the matrices it processes. The LSTM network's output is then passed through the second half of the U-Net, where a series of deconvolutional layers merges the feature matrices and outputs a single risk map.

The neural network model is evaluated through and trained by minimizing the Binary-Cross Entropy loss, defined in Equation 1, where $L(y)$ is the final loss function, $y$ is the set of labels (1 for yes wildfire, 0 for no wildfire), $p(y)$ is the probability of the given label, and $N$ is the total number of pixels on a prediction heatmap.

\begin{equation}
    L(y)=-\frac{1}{N}\sum_{0}^N y_i \cdot \log(p(y_i))+(1-y_i) \cdot log(1-p(y_i))
\end{equation}

The ULSTM, along with all other codes in this project, were written in Python, and the machine learning models were implemented using the PyTorch Libraries.

\subsection{A Proposed Personalized Early Warning System}

A personalized early warning system, which would give user-specific alerts and precautions given their circumstances and special conditions, would have a great impact for those in areas at risk of wildfires, especially those with disabilities or specific health conditions. A raw percentage value or recent trend of wildfire risk would have varied interpretations from different members of the same community, leading to mixed and likely uncoordinated responses. An early warning system would eliminate such confusion and provide actionable items for civilians and fire protection agenceis. Additionally, the personalized aspect of the system would allow more vulnerable populations, such as those with respiratory diseases, sensory impairments, or mobility disabilities, to better prepare in advance to plan the extra steps they would need to take.

Previous work on early-warning systems primarily focused on detecting, rather than predicting, wildfires. As reviewed earlier, existing prediction techniques are also limited in scope and/or accuracy. These techniques utilize regression and/or sensor threshold methods of assessing environmental conditions, which are less robust as compared to state-of-the-art neural network algorithms. Thus, implementing newer AI networks such as the ULSTM will improve the accuracy and reliability of early disaster warning systems. 

\begin{figure*}[thpb]
      \centering
      \label{figure-ews}

      \includegraphics[width = 6in]{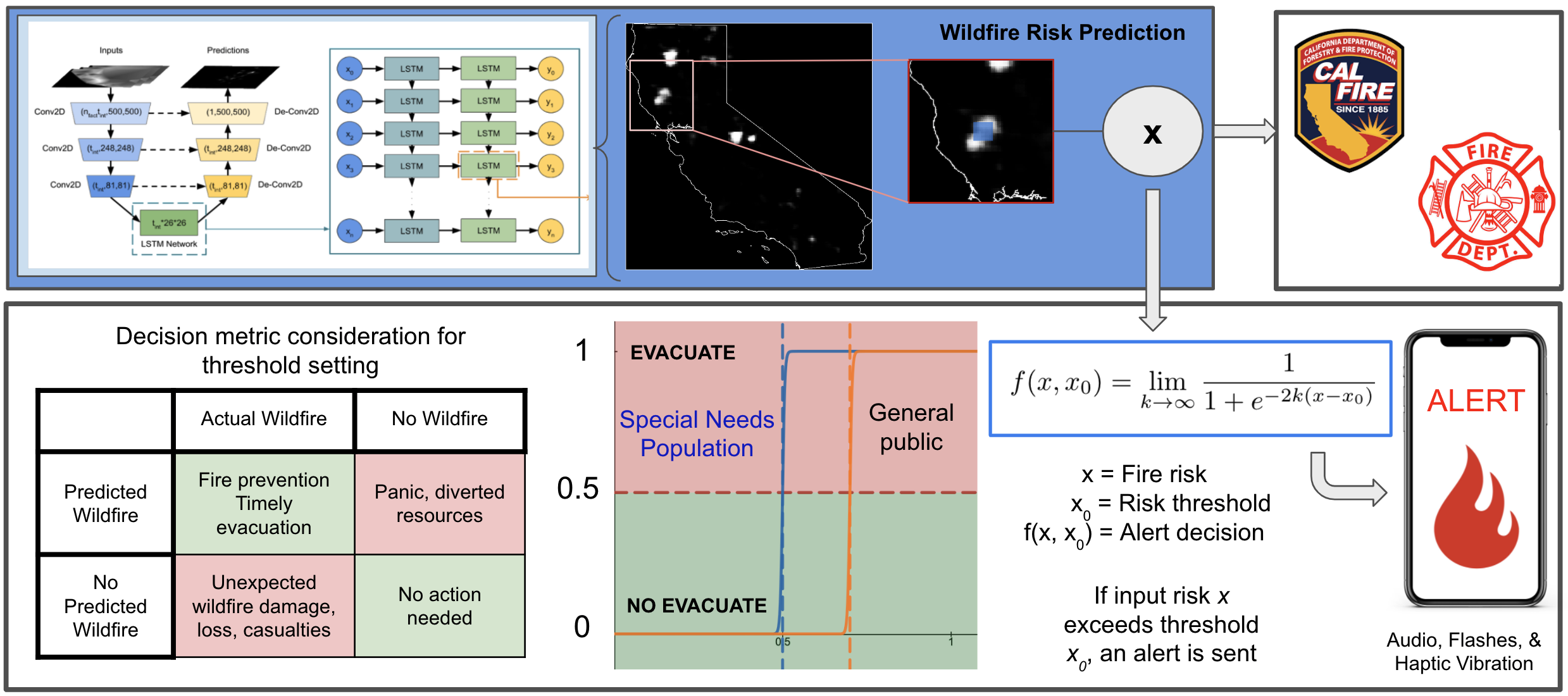}
      \caption{Diagram of the proposed early warning system based on wildfire prediction. The ULSTM uses environmental and meteorological data to generate risk maps over California, which is directly sent over to CalFire, fire departments, and other wildfire protection agencies. For regions of significantly high risk, impacted communities and fire protection agencies would be notified of potential fires and evacuation instructions. Individuals with specific circumstances, such as respiratory exacerbation risks, would have different warning criteria and content.}
\end{figure*}

\hyperref[figure-ews]{Figure 4} details the specifications of the proposed personalized early warning system. Utilizing the trained ULSTM neural network model to extract wildfire predictions, the warning system would identify the areas of high wildfire probabilities. The information would then be utilized by fire protection agencies as well as the early warning algorithm to decide if and to whom to send alerts to. For example, individuals with special conditions would receive warnings at different risk levels and proximity to predicted fires. Such adjustments would accommodate for the degree to which certain individuals are impacted by fires and their byproducts and/or are able to evacuate. These populations include those with:
\begin{enumerate}
  \item Health ailments that would make one more susceptible to environmental, physiological, or psychological stress,
  \item Mobility disabilities that would hampers one's ability to evacuate, and
  \item Mental conditions that would alter one's ability to comprehend an act upon alerts and warnings.
\end{enumerate}
For example, individuals with respiratory conditions such as chronic obstructive pulmonary diseases (COPD) or asthma, who are much more sensitive to smoke and pollution generated from wildfires, would be notified at a lower predicted risk level and larger distance from predicted fires than healthy individuals. Additionally, they would receive specific instructions to protect themselves, such as to stay indoors and wear masks.

Additionally, smoke and particulates from wildfires create a large amount of pollution which negatively affects air quality. Breathing air quality is one of the most crucial factors in human health; poor air quality can cause any person’s health to significantly deteriorate and is an increasingly important issue following the advent of rapid industrialization. Especially since their lungs are compromised due to inflammation, individuals with respiratory diseases such as COPD and asthma are extremely susceptible to exacerbation caused by bad air quality, leading to hospitalization. Based on the results of retrospective medical studies, a methodology to estimate the increase in exacerbation risk of respiratory conditions has previously been reported, using the four trailing indicators near the patient’s location [25]. Specifically, if a factor falls below the threshold standard, its contribution to the final risk percentage is zero; otherwise, it follows the formula as shown in Equation 2 [25].

\begin{equation*}
  \begin{array}{l}
    R = 1 \% \cdot \frac{([PM_{2.5}]-12 \frac{\mu g}{m^3})}{10. \frac{\mu g}{m^3}} + 0.8\%\cdot \frac{([PM_{10}]-7 \frac{\mu g}{m^3})}{10. \frac{\mu g}{m^3}}\\
    + 2 \% \cdot \frac{([NO_2]-101.23 \frac{\mu g}{m^3})}{10. \frac{\mu g}{m^3}} + 4.7\%\cdot \frac{68.0^{\circ}F-T_F}{1.8^{\circ}F}
  \end{array}
\end{equation*}

Based on the respiratory exacerbation risks determined for individuals with conditions such as COPD or asthma using the above formula, personalized alerts would be created and delivered for those at elevated levels of risk.

\section{\textbf{RESULTS}}

Using the comprehensive wildfire dataset described in the previous sections comprising the environmental and meteorological parameters, the proposed ULSTM neural network was trained and evaluated for accuracy of its wildfire predictions. 

The formula used to define the network's accuracy is given in Equation 3, where $t_{int}$ represents the interval of days being processed to make a prediction (7 days), $F[t']_{i,j}$ and $T[t']_{i,j}$ represents the value of the pixel at position ${i,j}$ of the predicted risk heat-map and true wildfire  heat-map (respectively) on day ${t'}$, and $t$ represents the total time duration.

\begin{equation*}
    \begin{array}{l}
        \hspace*{-1.6cm}
        \textbf{ACCURACY} = \\
        \hspace*{-1.6cm}

        \left(1 - \frac{1}{2}\sum_{t_0=0}^{\text{t}}\left(\frac{1}{t_{int}}\sum_{t_1=0}^{{t_{int}}}\sum_{i,j}f\left(F[t_0-t_1]_{i,j}\right) - \sum_{i,j}T[t_0]_{i,j}\right)^2\right) \cdot 100\%
    \end{array}\end{equation*}

In simple terms, the accuracy, which employs a modified version of mean-squared-error analysis, measures how close the network's predicted risk heat-map is to the true wildfire heat-map.

\begin{figure*}[thpb]
      \centering
      \label{figure-accuracies}
      \includegraphics[width = 6in]{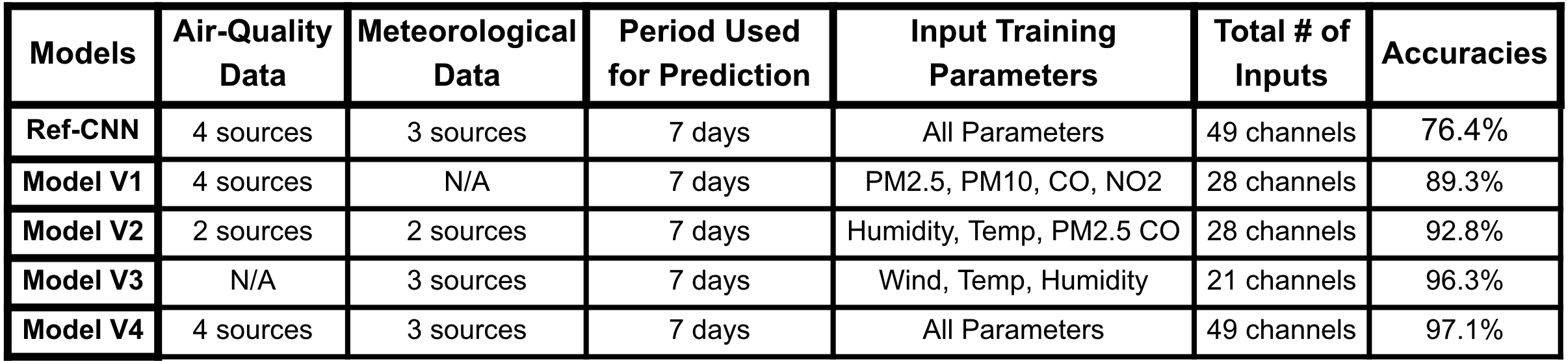}
      \caption{Table outlining the data sources and network inputs, period analyzed for prediction, and corresponding accuracies for five networks, including one benchmark LeNet-5 CNN and four variations of the ULSTM. These four variations are listed in increasing order of accuracy achieved, including Model V1 (trained only on trailing air-quality indicators), Model V2 (trained on a combination of two leading  indicators and two trailing indicators), Model V3 (trained only on leading meteorological data), and Model V4 (trained on all leading and trailing indicators).} 
\end{figure*}

To compare the magnitude to which the various leading and trailing indicators contribute to the network's final predictions, a series of four ULSTM networks, each with varying input shape (dependent on the number of indicators being analyzed) but otherwise sharing the same architectural features and specifications, were developed, trained, and evaluated. \hyperref[figure-accuracies]{Figure 5} shows the structures and results of these four models, which were each trained to analyze seven days worth of data to demarcate areas with high-risk of wildfire. Model V1 analyzes all four air-quality trailing indicators (PM$_{2.5}$, PM$_{10}$, CO, NO$_2$), achieving an accuracy of 89.3\%. Model V2 takes in two meteorological leading indicators and two air-quality trailing indicators (Humidity, Temperature, PM$_{2.5}$, CO), achieving an accuracy of 92.8\%. Model V3 relies on the three meteorological leading indicators (Wind Speed, Temperature, and Humidity) to achieve 96.3\% in accuracy. Finally, Model V4 takes in all seven total leading and trailing indicators (Wind Speed, Temperature, Humidity, PM$_{2.5}$, PM$_{10}$, CO, NO$_2$), achieving an overall accuracy of 97.1\%. 

The performance of the proposed ULSTM spatio-temporal neural network in predicting large wildfires was benchmarked against a traditional Convolutional Neural Network (CNN) architecture represented by the seminal LeNet-5 model [29]. The LeNet-5 architecture, consisting of three sets of convolutional layers followed by two fully connected layers, was trained over the same seven leading and trailing indicators and evaluated with the same dataset to compare the performances of the four ULSTM network variants. The CNN model achieved an accuracy of 76.4\%.

Thus, each of the four variants of the ULSTM network achieved significantly higher accuracy compared to the reference CNN network, as shown in Figure 5. These results illustrate that the ULSTM networks with their unique spatio-temporal architectures are better suited for the analysis of the wildfire dataset compared to the CNN. Indeed, the training data was inputted into the CNN without any regard towards or analysis of the time-series component of the leading and trailing indicators. As such, although the CNN should be adequately suited for extracting key spatial features within the input data for making wildfire predictions, it stumbles in recognizing the time-related patterns amongst images, leading to a lower accuracy for predicting wildfires based on environmental and meteorological factors.

Amongst the four trained ULSTM networks, the presence of specific leading and trailing indicators or lack thereof within the training data show clear trends within the accuracy. This pattern becomes most explicit when comparing Model V1 and Model V3. The former, trained only on trailing air-quality indicators, achieved 89.3\% accuracy, whereas the latter, trained only using the leading meteorological indicators, achieved 86.3\% accuracy. As the four ULSTM networks were all trained using the same hyperparameters and have the same overall structure, the difference in performance between these two version lies within the nature of their input data sources.

By nature, the three meteorological factors are considered leading indicators as they contain patterns or trends which can provide insight into the likelihood of wildfire in the near future as well as the volatility of the land to fuel a small fire into a severe conflagration, as such their data provides analysis before a wildfire. Meanwhile, the four air-quality factors are designated as trailing indicators since their more indicative trends appear after the initial outbreak of a fire, showing the emissions of wildfires that have just appeared and/or have not been detected or noticed yet, as such their data provides analysis during a wildfire. For the task of wildfire prediction, then, it is intuitively more beneficial to be able to find trends in leading indicators to foretell potential areas of wildfire days in advance rather than detect already-occurring wildfires. As such, it makes sense for Model V3 of the ULSTM to have a higher accuracy than that of Model V1, as the former performs wildfire prediction using leading indicators while the latter uses trailing indicators. Model V2 of the LSTM backs up this observation, showing that taking half of the leading indicators and half of the trailing indicators leads to an accuracy percentage right in between those of Model V1 and Model V3.

This is not to say, however, that one only needs leading indicators for wildfire prediction and can disregard trailing indicators altogether. The Model V4 variant of the ULSTM combines all leading and trailing indicators, outperforming even Model V3. Of course, taking in more data sources should allow the trained neural network to demonstrate an equal, if not greater, accuracy than Model V3. Why it performs noticeably better lies with the fact that the trailing indicators do provide some additional insights into wildfire activity. For areas where meteorological factors could not have predicted the onset of wildfires, such as deliberate and expedited arson, trailing indicators provide extra patterns and trends to gather information from them and utilize in computing the predictions.

\begin{figure*}[thpb]
      \centering
      \label{figure-samples}
      \includegraphics[width = 6in]{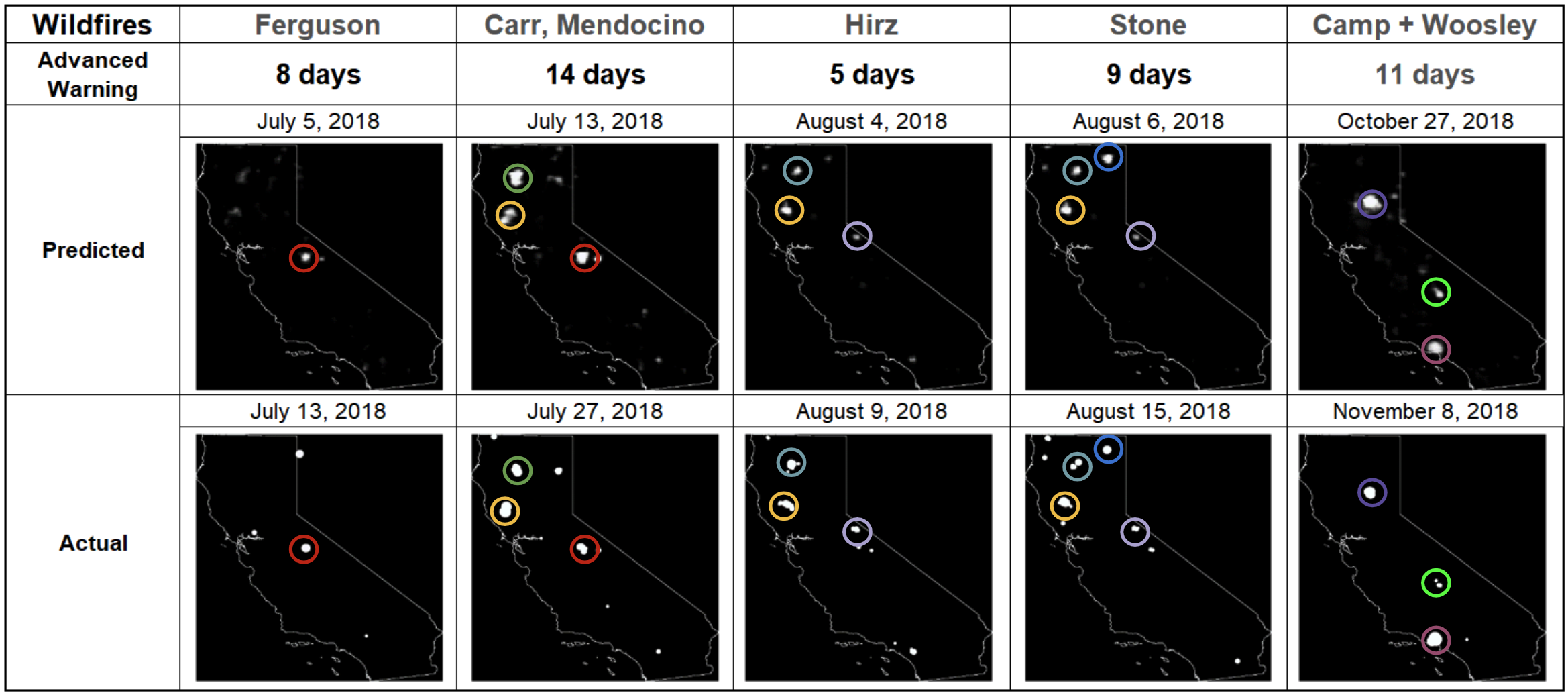}
      \caption{Table outlining instances of the ULSTM (Model V4, the best performing variant) predicting major wildfires, detailing wildfire names, dates and predicted and actual risk/wildfire heat-maps, and days in advance successful prediction was achieved. The ULSTM was able to successfully predict 7 of the 10 most devastating wildfires of the 2018 wildfire season (according to acreage burnt).}
\end{figure*}

\hyperref[figure-samples]{Figure 6} shows the wildfire prediction results generated by Model V4 of the ULSTM network compared with the actual occurrences of five large wildfires in California during the year 2018. Specifically, the first row of images shows the predicted wildfire risk-map output of the network, with each pixel value scaled between 0 and 1 for the purpose of visualization (with 0 being black representing the lowest risk and 1 being white representing the highest risk). Listed right above each predicted image is the day on which the wildfire prediction was generated (for example, July 5 would involve analyzing past 7 days of data, or indicators from June 28 to July 4). The second row of images represent the corresponding real-world instances of the wildfires, with the actual dates of occurrences listed above the images. Whereas the predicted image contains areas of high-likelihood of wildfires that would occur at a future date, the actual image shows the actual wildfire occurring in the same location days later (dates listed above the image). The topmost rows specify which wildfire had been identified in the predicted-actual pair as well as how many days before the actual fire did the successful prediction occur. Each circle on the map represents a region of high probability with multiple pixels being white, an area the network predicts will have high probability of wildfires in the near future. To differentiate wildfires and clarify if regions between images are the same wildfire, each high likelihood region is circled with a unique color across both the predicted and actual wildfire images. Specifically, for the ten largest wildfires in the 2018 California wildfire season, the red circle shows the Ferguson fire, yellow shows the Mendocino Complex fire, green shows Carr, teal shows Hirz, blue shows Stone, dark purple shows Camp, and magenta shows Woosley fires. The light purple and lime circles show smaller predicted wildfires. As Figure 6 highlights, seven of California's ten largest wildfires in 2018 was successfully predicted by Model V4 of the ULSTM network.

\begin{figure*}[thpb]
      \centering
      \label{figure-geological}
      \includegraphics[width = 6in]{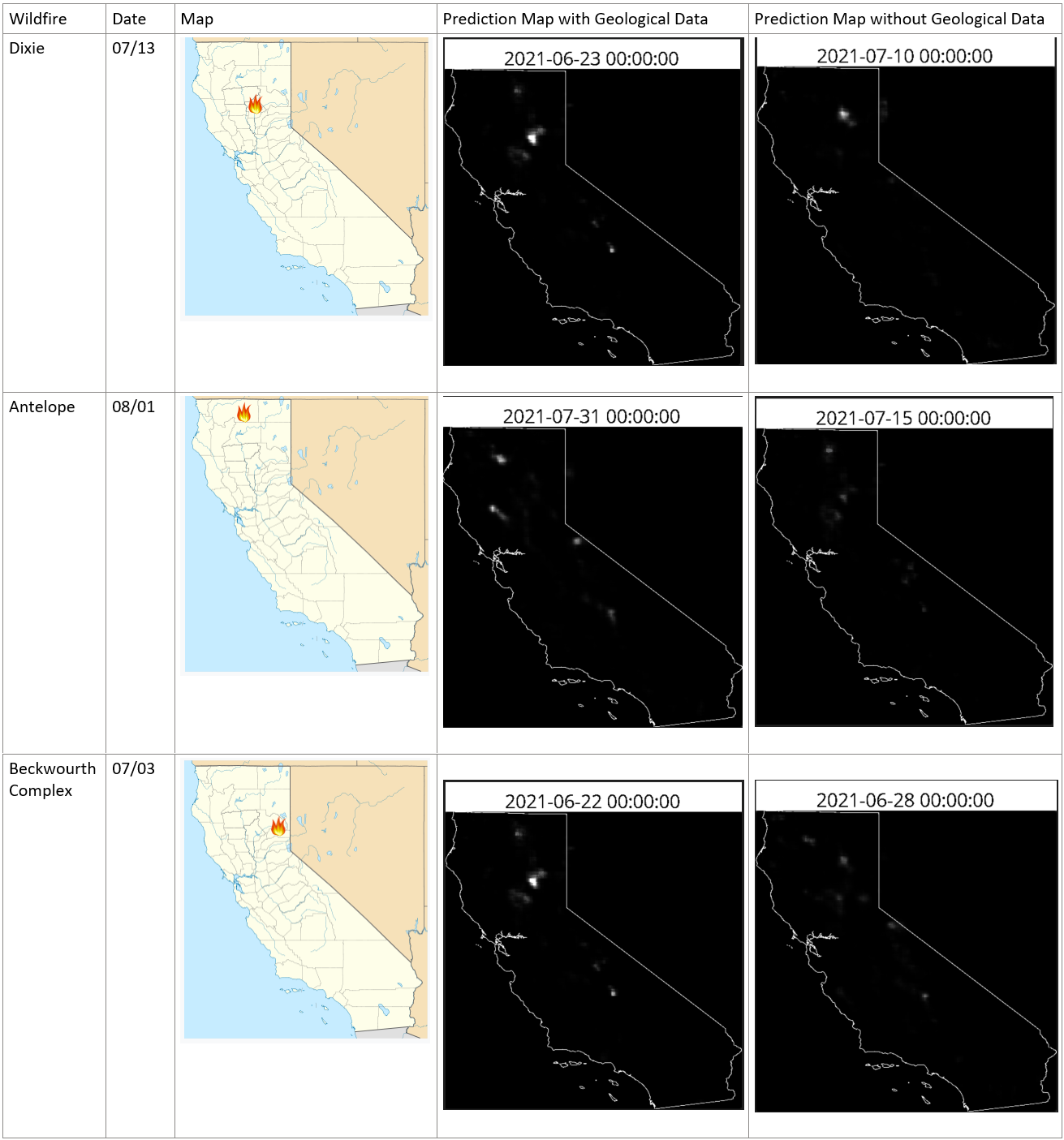}
      \caption{Table outlining the ULSTM neural network's performance trained with and without geological data with prediction maps in the left and right columns, respectively, analyzed over California's 2021 wildfire season. The ULSTM with geological data shows higher confidence and more accurate predicted location for impending wildfires.}
\end{figure*}

The aforementioned results detail the performance of the ULSTM networks trained solely on the leading meteorological and trailing air-quality indicators. Further enhancements and optimizations were performed on Model V4 including geological information. \hyperref[figure-geological]{Figure 7} details the performance comparison of the ULSTM network trained with and without the geological factor data (including NDVI vegetation data and terrain elevation data) analyzed over the largest fires of the 2021 wildfire season. The items in the first three columns include the name of the wildfire in question, the date of its initial outbreak, and its corresponding location on the map of California. The last two columns show the prediction maps generated by the ULSTM network trained with and without geological data, respectively. Scaling has not been applied on these images to allow for accurate analysis of the differences between the maps and the impact of the geological data. The prediction maps generated by the ULSTM network trained with geological data shows higher confidence (brighter pixel value) when predicting the Antelope and Beckwourth Complex fires. Additionally, it appears to have better prediction location for the Dixie and Beckwourth Complex fires. The inclusion of the geological data provides the ULSTM network with additional information regarding the land over which it predicts wildfires, giving it insights into which areas of land are more or less likely to host a wildfire outbreak. Hence, the ULSTM network version with geological data can make more informed decisions as to the specific details of impending wildfires, leading to higher confidence and closer proximity for predicted wildfires.


\section{\textbf{CONCLUSIONS}}

In summary, a novel spatio-temporal machine learning framework, based on a U-Conv-LSTM neural network architecture, was developed for accurate prediction of wildfires. A comprehensive wildfire database was created, largest of its kind, containing environmental and meteorological indicators, geological information, and wildfire data spanning from 2012 through 2018. This database was successfully used to train and evaluate the ULSTM neural network, achieving up to 97\% prediction accuracy for up to 2 weeks in advance of the largest wildfires of the 2018 wildfire season in California. A proposed early warning system was developed to translate the wildfire risk assessment data obtained from the ULSTM model to create personalized alerts and warning for at-risk individuals in the affected areas, such as those with respiratory exacerbation risks, as well as data for wildfire protection agencies to use.

Overall, the unique spatio-temporal neural network architecture demonstrated relatively high accuracy in pinpointing the location and severity of the largest wildfires of California. Deployment of such a technique would allow fire protection agencies and firefighters to divert resources to high-risk areas, confront fires early, and mitigate potential damages to the economy, infrastructure, and quality of life. This solution, however, should not solely dictate the resulting response towards potential wildfires; rather, it is meant as an assistive tool for helping make better preparations and decisions regarding the wildfire season, such as stationing firefighters, warning the public, and facilitating targeted evacuations in advance. The proposed personalized early warning system can potentially act as a communication bridge between firefighting agencies and the public: while wildfire management officials can educate the public on the causes and mechanisms of wildfires, suggest certain practices to reduce the likelihood of local wildfires, and warn of upcoming procedures such as prescribed burns which could release some air pollution, the public too can directly get into contact and potentially provide alerts of wildfire outbreaks that would enable firefighters extra time to put out the wildfire early before it exacerbates significantly.

In the future, expanding the scope of this wildfire prediction solution to the entire U.S. would allow this model to have a much larger impact. Although California's wildfire seasons are and likely will continue to be relatively more damaging, larger attention and funding towards wildfire suppression as well as less foliage to burn through means that California's wildfire seasons might become less of an issue towards the near future. This however, does not warrant any less importance or significance placed on the statewide epidemic of wildfires, which has destroyed plenty of infrastructure and lives, wreaked havoc on California's forestry and atmosphere, and sucked out billions of dollars from California's budget which would otherwise be used to advance the society in a variety of ways. The issue of wildfires does not just pertain to California, however, as climate change continues to impact the nation and the rest of the world as a whole, with globally increasing temperatures leading to drier foliage and higher likelihood of wildfires igniting and burning out of control. Wildfires in states with a less prepared fire protection agencies could impart significant damages. Thus, a nationwide fire prediction tool could help save lives and prevent or limit damages with proactive intervention in advance of the impending disasters.

\section*{\textbf{ACKNOWLEDGMENT}}

The author would like to express sincere gratitude to C. Spenner for helping shape the project and providing insightful advice and S. Singla of UC Riverside for her generous guidance in navigating and utilizing the WildfireDB dataset.


\begin{thebibliography}{26}
\bibitem{c1} K Hoover, LA Hanson, “Wildfire Statistics,” Congressional Research Service, Washington, D.C., U.S., IF10244, Aug. 1, 2022. Accessed: Aug. 10, 2022. [Online]. Available: https://sgp.fas.org/crs/misc/IF10244.pdf
\label{c1}
\bibitem{c2} TW Porter, W Crowfoot, G Newsom, “Wildfire Activity Statistics,” CALFIRE, Sacramento, California, 2020. Accessed: Aug. 10, 2022. [Online]. Available: https://www.fire.ca.gov/media/0fdfj2h1/2020\_redbook\_final.pdf
\label{c2}
\bibitem{c3} P Folger, “Drought in the United States: Causes and Current Understanding,” Congressional Research Service, Washington, D.C., U.S., R43407, Nov. 9, 2017. Accessed: Aug. 10, 2022. [Online]. Available: https://sgp.fas.org/crs/misc/R43407.pdf
\label{c3}
\bibitem{c4} PA Ullrich, Z Xu, AM Rhoades, MD Dettinger, JF Mount, AD Jones, and P Vahmani. California's drought of the future: A midcentury recreation of the exceptional conditions of 2012–2017. Earth's Future, 6, 1568–1587 (2018). https://doi.org/10.1029/2018EF001007
\label{c4}
\bibitem{c5} R Miller. Prescribed Burns in California: A Historical Case Study of the Integration of Scientific Research and Policy. Fire, 3, 44 (2020). https://doi.org/10.3390/fire3030044
\label{c5}
\bibitem{c6} JE Keeley, AD Syphard. Large California wildfires: 2020 fires in historical context. fire ecol 17, 22 (2021). https://doi.org/10.1186/s42408-021-00110-7
\label{c6}
\bibitem{c7} P Jain, SCP Coogan, SG Subramanian, M Crowley, S Taylor, and MD Flannigan, “A review of machine learning applications in wildfire science and management,” Env. Rev. 28, 4, 478-505 (2020). https://doi.org/10.1139/er-2020-0019
\label{c7}
\bibitem{c8} Y Ban, P Zhang, A Nascetti, AR Bevington, and MA Wulder, “Near Real-Time Wildfire Progression Monitoring with Sentinel-1 SAR Time Series and Deep Learning,” Sci. Rep. 10, 1322 (2020). https://doi.org/10.1038/s41598-019-56967-x
\label{c8}
\bibitem{c9} A Molina-Pico, D Cuesta-Frau, A Araujo, J Alejandre, and A Rozas, “Forest Monitoring and Wildland Early Fire Detection by a Hierarchical Wireless Sensor Network,” Sensors Env. Monitoring 2016, 8325845 (2016). https://doi.org/10.1155/2016/8325845
\label{c9}
\bibitem{c10} S Kelleher, C Quinn, D Miller-Lionberg, and J Volckens, “A low-cost particulate matter (PM2.5) monitor for wildland fire smoke,” Atmos. Meas. Tech., 11, 1087–1097 (2018). https://doi.org/10.5194/amt-11-1087-2018
\label{c10}
\bibitem{c11} N Varela, JL Díaz-Martinez, A Ospino, NAL Zelaya, “Wireless sensor network for forest fire detection,” Proc. Comp. Sci., 175, 435–440 (2020). https://doi.org/10.1016/j.procs.2020.07.061
\label{c11}
\bibitem{c12} YO Sayad, H Mousannif, Hassan Al Moatassime, “Predictive modeling of wildfires: A new dataset and machine learning approach,” Fire Safety Journ. 104, 130-146 (2019). https://doi.org/10.1016/j.firesaf.2019.01.006
\label{c12}
\bibitem{c13} JC Liu, LJ Mickley, MP Sulprizio, F Dominici, X Yue, K Ebisu, GB Anderson, RFA Khan, MA Bravo, and ML Bell, "Particulate Air Pollution from Wildfires in the Western US under Climate Change," Clim Change. 138, 3, 655-666 (2016). https://doi.org/10.1007/s10584-016-1762-6.
\label{c13}
\bibitem{c14} MM Bela, N Kille, SA McKeen, J Romero-Alvarez, R Ahmadov, E James et al.  Quantifying carbon monoxide emissions on the scale of large wildfires. Geophys. Res. Let., 49, e2021GL095831 (2022). https://doi.org/10.1029/2021GL095831
\label{c14}
\bibitem{c15} D Griffin et al., "Biomass burning nitrogen dioxide emissions derived from space with TROPOMI: methodology and validation," Atmos. Meas. Tech. 14, 12, 7929–7957 (2021). https://doi.org/10.5194/amt-14-7929-2021
\label{c15}
\bibitem{c16} ML Ramsey, M Murphy, J Diaz, “The Camp Fire Public Report,” Butte County District Attorney, Oroville, California, 2020. Accessed: Aug. 10, 2022. [Online]. Available: https://www.buttecounty.net/Portals/30/CFReport/PGE-THE-CAMP-FIRE-PUBLIC-REPORT.pdf
\label{c16}
\bibitem{c17} "Air Data: Air Quality Data Collected at Outdoor Monitors Across the US," United States Environmental Protection Agency, Washingdon, D.C., U.S., Nov. 2, 2021. Accessed: Aug. 10, 2022. [Online]. Available: https://www.epa.gov/outdoor-air-quality-data
\label{c17}
\bibitem{c18} S. Mansoor et. al., "Elevation in wildfire frequencies with respect to the climate change", Journ. of Env. Manag. 301, 113769 (2022). https://doi.org/10.1016/j.jenvman.2021.113769
\label{c18}
\bibitem{c19} S Huang, L Tang, JP Hupy et al., "A commentary review on the use of normalized difference vegetation index (NDVI) in the era of popular remote sensing," J. For. Res. 32, 1–6 (2021). https://doi.org/10.1007/s11676-020-01155-1
\label{c19}
\bibitem{c20} S Singla, T Diao, A Mukhopadhyay, A Eldawy, R. Shachter, and M Kochenderfer, "WildfireDB: A Spatio-Temporal Dataset CombiningWildfire Occurrence with Relevant Covariates," AI for Earth Sciences Workshop at NeurIPS 2020
\label{c20}
\bibitem{c21} S Deshmukh, B Raj and R Singh, "Multi-Task Learning for Interpretable Weakly Labelled Sound Event Detection," (2020). https://doi.org/10.48550/arXiv.2008.07085
\label{c21}
\bibitem{c22} CM Kim, KH Kim, YS Lee, K Chung and RC Park, "Real-Time Streaming Image Based PP2LFA-CRNN Model for Facial Sentiment Analysis," IEEE Access, 8, 199586-199602 (2020). https://doi.org/10.1109/ACCESS.2020.3034319
\bibitem{c23} K Sanjeevan and T Hung, "PyTorch Audio Classification: Urban Sounds," (2020). https://github.com/ksanjeevan/crnn-audio-classification
\label{c23}
\bibitem{c24} Y. Xu, Q. Kong, W. Wang, and M. D. Plumbley, "Large-scale weakly supervised audio classification using gated convolutional neural network," CoRR (2017). https://doi.org/10.48550/arXiv.1710.00343
\label{c24}
\bibitem{c25} R Bhowmik, "A Personalized Respiratory Disease Exacerbation Prediction Technique Based on a Novel Spatio-Temporal Machine Learning Architecture and Local Environmental Sensor Networks," Electronics (2022). Accepted for publication on August 7
\label{c25}
\bibitem{c26} O Ronneberger, P Fischer, and T Brox, "U-Net: Convolutional Networks for Biomedical Image Segmentation", arXiv (2015). https://doi.org/10.48550/arXiv.1505.04597
\label{c26}
\bibitem{c27} E Shelhamer, J Long, T Darrell, "Fully Convolutional Networks for Semantic Segmentation", arXiv (2016). https://doi.org/10.48550/arXiv.1605.06211
\label{c27}
\bibitem{c28} S Hochreiter, and J Schmidhuber, "Long Short-Term Memory," Neur. Comp. 9, 8, 1735-1780 (1997). https://doi.org/10.5194/amt-14-7929-2021
\label{c28}
\bibitem{c29} Y LeCun, L Bottou, Y Bengio, and P Haffner, "Gradient-Based Learning Applied to Document Recognition," Proceedings of the IEEE. 86 (11), 2278-2324 (1998). https://doi.org/10.1109/5.726791
\label{c29}
\end{thebibliography}
\end{document}